\documentclass[11pt,a4paper]{article}
\usepackage{times,latexsym}
\usepackage{url}
\usepackage[T1]{fontenc}

  \usepackage[acceptedWithA]{tacl2021v1}
\usepackage{tacl2021v1}

\usepackage{xspace,mfirstuc,tabulary}
\usepackage[table]{xcolor}
\usepackage{times}
\usepackage{latexsym}
\usepackage{graphicx}
\usepackage{tikz}
\usetikzlibrary{shapes.geometric, arrows, positioning}
\usepackage{amsmath}
\usepackage{multirow}
\usepackage{booktabs}
\usepackage{inconsolata}
\usepackage{enumitem}
\usepackage{comment}

\newif\iftaclinstructions
\taclinstructionsfalse 
\iftaclinstructions
\renewcommand{\confidential}{}
\renewcommand{\anonsubtext}{(No author info supplied here, for consistency with
TACL-submission anonymization requirements)}
\newcommand{\instr}
\fi

\iftaclpubformat 

\else

\fi

\newif\ifcomments
\commentstrue
\ifcomments
    \providecommand\rotem[1]{\textcolor{blue}{[Rotem: #1]}}
    \providecommand\adir[1]{\textcolor{magenta}{[Adir: #1]}}
    \providecommand\zohar[1]{\textcolor{green}{[Zohar: #1]}}
    \providecommand\yuval[1]{\textcolor{orange}{[Yuval: #1]}}
    \providecommand\elad[1]{\textcolor{pink}{[Elad: #1]}} 
    \providecommand\oshrat[1]{\textcolor{brown}{[Oshrat: #1]}}
    
\else
    \providecommand{\rotem}[1]{}
    \providecommand{\adir}[1]{}
    \providecommand{\zohar}[1]{}
    \providecommand{\yuval}[1]{}
    \providecomand{\oshrat}[1]{}
    \providecommand{\elad}[1]{}    
\fi

\title{\emph{Deep Persona}: A Psychologically Grounded Architecture and Evaluation Framework for Role-Playing Agents and Simulations}

\author{Rotem Dror$^1$, Zohar Elyoseph$^2$, Yuval Haber$^3$, Elad Refoua$^4$, Oshrat Ayalon$^1$, Adir Solomon$^1$\\ 
\ \\
$^1$Faculty of Computer and Information Science, University of Haifa, Israel\\
\texttt{rdror@is.haifa.ac.il}\\
$^2$School of Therapy, Counseling and Human Development, University of Haifa, Israel\\
$^3$The PhD Program of Hermeneutics and Culture, Interdisciplinary Studies Unit, Bar-Ilan University, Israel\\
$^4$Department of Psychology, Bar-Ilan University, Israel\\
}

\date{}

\begin{document}
\maketitle
\begin{abstract}
    Existing approaches to persona simulation with Large Language Models (LLMs) mostly rely on shallow character descriptions that fail to sustain coherent character behavior across extended interactions. We introduce \emph{Deep Persona}, a psychologically grounded, three-layered architecture that organizes personas into hierarchical levels of observable expression, latent beliefs, and core motivational drives, for constructing highly convincing role-playing agents. Governed by the principles of scripted determinism and bounded agency, the architecture restricts the model to a reactive engine guided by a structured internal script. We further propose a reference-free evaluation framework that benchmarks dialogue naturalness against empirical human distributions using established psychological clinical instruments and adversarial stress-tests. Empirical evaluation reveals that while LLMs achieve high pragmatic fluency, they exhibit systematic limitations in emotional expression and joint attention. In addition, we present a case study of two \emph{Deep Personas} and evaluate them using the proposed framework, demonstrating that structured personas can produce interactions that more closely align with human conversational behavior.
\end{abstract}

\section{Introduction}\label{sec:intro}

LLMs are increasingly used as interactive agents capable of simulating real people across a wide range of domains~\cite{tseng2024two}. These systems support applications such as conversational assistants, educational tools, entertainment platforms, and training environments, where models are expected to adopt specific identities and engage in multi-turn interactions. Recent work has demonstrated the ability of LLMs to perform role-playing in conversational settings \cite{tao2024chatgpt, wang2024rolellm, zhou2025characterbench}. In particular, LLM-based agents are increasingly explored in mental health and clinical training, where simulated interactions support the education, evaluation, and skill development of therapists \cite{lawrence2024opportunities, hua2025scoping, elyoseph2026effectiveness}. These applications place strong demands on the realism, consistency, and stability of simulated personas.

\begin{figure}
    \centering
    \includegraphics[width=\linewidth]{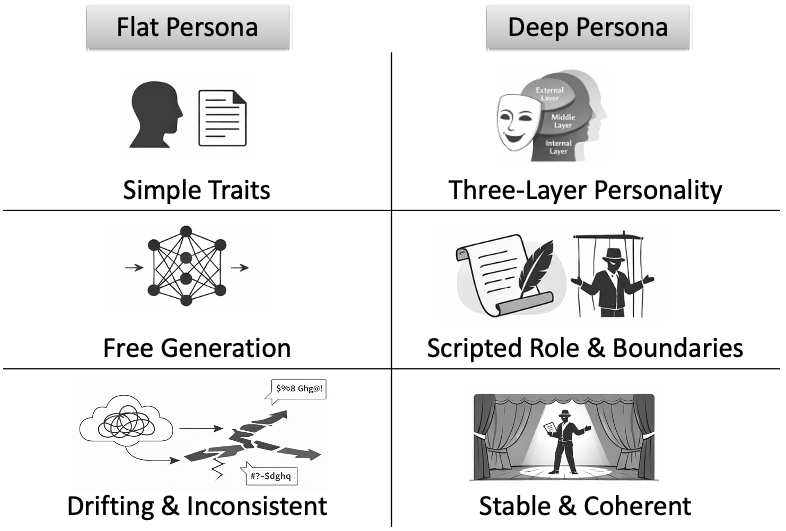}
    \caption{An illustration of the differences between a flat persona and the \emph{Deep Persona} framework.}
    \label{fig:illustration}
\end{figure}

Despite this progress, current approaches to persona modeling remain fundamentally limited. In most existing work, personas are defined through flat representations, typically consisting of short prompts that describe surface-level attributes such as personality traits or assigned roles. While such approaches can guide local response generation, they often fail to maintain coherence across extended interactions~\cite{li2025hello}, leaving models prone to hallucinations, inconsistencies, and gradual drift from the intended character \cite{wang2024rolellm, zhou2025characterbench}. 

This work offers two main contributions. First, we propose a novel methodology for constructing personas that introduces psychological depth and internal structure beyond flat trait descriptions. Our approach, termed \emph{Deep Persona}, models a character not only through observable behavior but also through underlying motivations, beliefs, and constraints that govern response generation. By explicitly encoding the factors that shape responses, the resulting agent becomes more consistent, interpretable, and human-like. A strict and comprehensive specification further constrains the agent’s role, interaction boundaries, and permissible actions, reducing hallucinations and the risk of breaking character. In this paper, we include guidelines for eliciting character specifications from domain experts, organizing information into layered structures, and defining interaction dynamics to support long-term coherence.



Second, we introduce a unique evaluation framework for assessing the realism and naturalness of persona utterances in dialogue. A central component of this framework is inspired by the Autism Diagnostic Observation Schedule (ADOS) \cite{lord2000autism}, a clinical instrument used to assess social and communicative behavior. We adapt key dimensions of this test into automated metrics capturing pragmatic fluency, joint attention, emotional congruence, and affective diversity. These measures provide a reference-free approach to evaluating the naturalness of LLM-generated dialogue throughout an interaction. In addition, we introduce the \textbf{Dialogue Naturalness Score (DNS)}, a statistical measure that quantifies the similarity between model-generated and human dialogue distributions. This is achieved by comparing the model’s scoring profile with human baselines using the Mahalanobis distance, enabling formal hypothesis testing to determine whether the LLM-generated dialogue is statistically indistinguishable from human interaction.


We demonstrate the application of our proposed evaluation framework on two human–human dialogue datasets \cite{li-etal-2017-dailydialog, bird2024generative} and three human–LLM datasets \cite{tao2024chatgpt, finch2023don, bird2024generative}. The results reveals systematic differences between human and LLM-generated dialogue, particularly in emotional calibration and contextual coherence. We further apply the full framework, including an embodied expression component that, to the best of our knowledge, has not been previously examined or implemented in this context, to two \emph{Deep Persona} simulations in a suicide risk assessment training setting and a parental mentalization scenario. These case studies provide the first direct evaluation of the proposed architecture, with the resulting interactions achieving higher DNS scores than the examined human–LLM datasets.



\section{Related Work}\label{sec:background}

\subsection{Persona Modeling with LLMs}
Recent work has explored the capacity of LLMs to simulate personas and perform role-playing in dialogue~\cite{tao2024chatgpt, wang2024characteristic}. Existing approaches for persona construction primarily rely on prompt-based role assignment, instructing LLMs to adopt a specific identity, personality trait, or narrative background to personalize conversational agents~\cite{zhang2018personalizing}, or to simulate specific human subpopulations for social and behavioral research~\cite{park2023generative}. 
These efforts are accompanied by evaluation frameworks that assess personas' ability to sustain stylistic fidelity, character consistency, and behavioral coherence across multiple dialogue turns~\cite{wang2024rolellm, zhou2025characterbench, ha2024clochat}. These advancements are further supported by empirical resources, including datasets like PersonaChat~\citep{zhang2018personalizing} and corpora of real-world conversations \citep{shuster2021dialogue, tao2024chatgpt}.


While these studies demonstrate that LLMs can simulate human interactions, most existing approaches represent personas superficially. As a result, they often focus on evaluating stylistic consistency rather than the realism and psychological depth of the persona.

\subsection{Evaluating Human-Likeness of LLM Dialogue}
Another growing body of research focuses on evaluating the human-likeness of LLM-generated text. Early work addressed the problem of detecting machine-generated text by identifying statistical linguistic deviations using measures such as token-likelihood distributions, perplexity, or entropy~\citep{beresneva2016computer, gehrmann2019gltr, mitchell2023detectgpt, su2023detectllm}. More recent approaches frame detection as a supervised classification task, fine-tuning LLMs to distinguish between human- and machine-generated text \citep{wang2023seqxgpt, liu2023coco}. 

Closely related to our work are studies that evaluate the linguistic quality and human-likeness of LLM outputs, rather than simply detecting their origin. For example, \citet{lu2025evaluating} compared LLM-generated and human-authored responses in role-play scenarios using evaluation dimensions derived from \citet{mehri2020usr}, including naturalness, contextual fluency, and overall response quality. Similarly,~\citet{duan2024hlb} proposed a Human-Likeness Benchmark (HLB) that evaluates models across psycholinguistic dimensions such as lexical choice, syntax, semantics, and discourse structure. 

While these approaches provide valuable information on the linguistic similarity between human and model-generated text, they primarily evaluate isolated responses or model-level capabilities. They do not explicitly address the evaluation of structured personas operating within interactive simulations, where emotional expression, behavioral consistency, and dialogue dynamics are central to the experience.

\subsection{LLM-Based Simulation in Mental Health}
High-fidelity persona simulation is increasingly vital for mental health training amid global professional shortages~\citep{lawrence2024opportunities, hua2025scoping}. By acting as simulated patients, LLMs allow clinicians to practice therapeutic skills in reproducible and controlled environments, with demonstrated promise in applications like suicide risk assessment and crisis response~\citep{elyoseph2026effectiveness, zhao2025effect, haber2025validating}. While LLMs lack the accountability required to replace human therapists~\citep{moore2025expressing}, they are highly suited for simulating patients. However, an effective simulation demands agents capable of maintaining psychologically coherent and reliable personas across extended interactions, which still constitutes a challenge for existing frameworks.





In summary, previous work has explored persona prompting techniques, benchmarks for role-playing ability, methods for evaluating human-likeness of generated language, and applications of LLM-based agents in training environments. Nevertheless, existing approaches typically define personas superficially and evaluate them using general linguistic metrics. 
In this work, we address these limitations by introducing a psychologically grounded architecture for constructing \emph{Deep Personas}, along with an evaluation framework inspired by psychological models measuring behavioral coherence, emotional congruence, and interactional naturalness of LLM personas.

\section{Foundational Design Principles}\label{sec:assumptions}

Our methodology for constructing a \emph{Deep Persona} relies on several assumptions regarding the capabilities and limitations of LLMs in interactive simulations. These assumptions inform both the design of the persona architecture and the prompting strategies used to implement it.

\subsection{Principle of Scripted Determinism}
A defining characteristic of our methodology is the deliberate refusal to rely on a model's inherent capabilities, ``intelligence,'' or training data as a basis for behavioral consistency. Instead, we assume that any behavior not explicitly encoded in the system prompt will inevitably degrade over time due to stochastic drift~\cite{liu2024lost}. This approach treats the LLM not as an autonomous agent with discretionary judgment, but as a stochastic engine that requires a rigid and elaborate set of rules to function appropriately.

Consequently, the prompt is designed as a detailed specification of the character and its interaction constraints. Rather than delegating responsibility for the conversational trajectory to the model, the prompt defines the boundaries within which the model operates. In this view, the prompt engineer assumes the role of a director who structures the interaction, while the LLM functions as an actor executing a predefined role.

\subsection{Principle of Bounded Agency}
LLMs struggle to maintain coherent behavior in roles that demand proactive leadership or open-ended expertise over long interactions (e.g., a therapist or scientist). Such roles require handling an unbounded range of user inputs while maintaining a consistent long-term strategy. To improve reliability, we propose restricting the model to roles with reactive and bounded agency. Instead of leading the interaction, the agent participates in a predefined scenario and responds to the user within a constrained narrative context. By explicitly defining the limits of the agent's world (the ``script'') and its responsibilities, we reduce hallucination, role drift, and character break.

\subsection{The Three Layers of Personality}\label{subsec:three_layers}
We assume that credible personas require an internal structure rather than a list of attributes. Inspired by psychological models of personality~\citep{freud1961ego, mcadams2006new}, we represent the persona as a multi-tiered information system that separates observable behavior (i.e., what text the agent generates) from underlying motivations (i.e., what are the reasons for generating a certain response). Importantly, we do not claim that this formulation instantiates genuine psychological constructs such as an ``unconscious mind.'' Rather, these layers serve as a design abstraction that enables the persona to exhibit behavior consistent with deeper internal states.

\begin{itemize}
\item \textbf{The External Layer (Conscious):} The publicly expressed identity of the persona, encompassing observable behavior, communication style, and emotional tone. This corresponds to the standard persona prompt used in prior work on persona modeling (e.g., \citet{tao2024chatgpt,hu2024quantifying}). 
\item \textbf{Middle Layer (Pre-Conscious):} Internal beliefs, attitudes, and contextual information that may influence responses. Content encoded in this layer is revealed only if the interaction evolves in a specific direction or if the user navigates the conversation to trigger it (e.g., triggers for withdrawal or deflection).
\item \textbf{Internal Layer (Unconscious):} This is the most critical and counter-intuitive layer. It consists of the persona's core motivations, hidden constraints, and psychological drives that guide behavior but are never explicitly verbalized.
\end{itemize}

\subsection{Persona Dynamics}
We assume that a realistic interaction requires the persona to evolve over time. Static characters that respond identically throughout the dialogue fail to produce believable simulations. Therefore, the interaction is structured into predefined stages representing distinct psychological or situational states (e.g., guarded, cooperative, reflective). Transitions between stages are governed by explicit triggers, including turn count or specific user behaviors. 

\section{Process of Persona Construction}\label{sec:construction}

This section describes the process of translating the design principles described above into a functional \emph{Deep Persona} by (1) eliciting the psychological and narrative structure of the character from a domain expert, and (2) encoding this specification into a structured prompt.

\subsection{Expert Interview: Knowledge Elicitation}

Persona construction begins with a structured interview with a domain expert who defines the character to be simulated. This interview is based on a proprietary methodology implemented by an AI agent within the Cesura.ai system, a platform for designing and operating simulation-based environments used to train and assess soft skills across organizational and institutional settings.
The purpose of the interview is to define the psychological narrative and interactional information required to construct a coherent persona. Rather than providing only a short role description (e.g., ``a depressed patient'') or general demographic characteristics (e.g., age, gender), the expert is asked to specify the character’s internal world in detail. The interview therefore collects information about the persona’s background, communication style, motivations, emotional patterns, and expected behavioral trajectory during the interaction. This procedure resembles a guided imaginative exercise in which the expert is asked to construct a detailed representation of the character. The process is structured through questions such as: What is the character’s personal history and social context? In what concrete situation does the interaction begin? What motivations shape the character’s behavior? 
Table~\ref{tab:interview_to_prompt} presents a full list of recommended content.

\begin{table*}[t]
\centering
\small
\begin{tabulary}{\linewidth}{p{0.2\linewidth} L L}
\hline
\textbf{Interview Dimension} &
\textbf{Elicited Content} &
\textbf{Prompt Component} \\
\hline
Narrative Background &
Personal history, social context &
Narrative Background and Motivations \\
\hline
Initial Context &
Entry setting (time/place) &
Interaction Structure: Initialization \\
\hline
Emotional Tone &
Affect, stance, typical mood &
External Layer \\
\hline
Communication Patterns &
Vocabulary, sentence length, questioning style &
Interaction Structure: Parameters \\
\hline
Conscious Goals &
Stated desires, explicit complaints, surface motivations &
External Layer \\
\hline
Semi-conscious Material &
Information revealed only after trust or progression &
Middle Layer \\
\hline
Unconscious Drives &
Conflicts, fears, avoidance patterns &
Internal Layer\\
\hline
Psychological Needs &
Attachment, validation, autonomy, control &
Internal Layer \\
\hline
Resistance Patterns &
Triggers for withdrawal, deflection, or shutdown &
Middle Layer \\
\hline
Evolution &
Expected change trajectory across interaction &
Control and Logic Module \\
\hline
Triggers for Change &
User behaviors that advance or stall progress &
Control and Logic Module \\
\hline
Ethical and Behavioral Boundaries &
Actions and disclosures that are prohibited &
Control and Logic Module \\
\hline
\end{tabulary}
\caption{Mapping interview dimensions to prompt specification components in the \emph{Deep Persona} methodology.}
\label{tab:interview_to_prompt}
\end{table*}

Importantly, the interview captures information across different levels of psychological accessibility. The expert specifies not only what the persona would openly communicate, but also hidden motivations, internal conflicts, and contextual information that influence behavior without being explicitly verbalized. 
Another practical implication of this is that the interview and subsequent prompt construction are ideally conducted in the native language of the persona designer, or in the language in which the persona is expected to communicate \cite{elyoseph2026effectiveness}. Linguistic features, such as register, rhythm, idiomatic expressions, and pragmatic norms, are treated as integral components of the persona. 

\subsection{Persona Prompt Architecture}\label{sec:body_lang}
Once the persona specification has been defined by the expert, it is translated into a structured prompt that functions as a specification document of the character’s internal world and interaction constraints.
Previous work on AI-based simulations has shown that long-term role consistency cannot be reliably achieved by instructing the model how to behave; instead, behavior must emerge from a well-defined internal context that constrains the model’s responses \citep{elyoseph2026effectiveness, levkovich2025step, kariv2025ai, haber2025validating}. Our approach therefore follows a closed-world design paradigm in which the persona’s internal reality is fully specified and the model generates responses consistent with it. The prompt architecture is organized into functional modules that jointly control the agent’s behavior during the interaction as described below.

\paragraph{A Short Introduction}
The opening section of the prompt specifies the type of interaction to be conducted and the high-level objectives governing it. This section informs the agent of the role it is expected to enact and the functional purpose of the exchange (e.g., guiding decision-making or providing structured feedback). 
An example of an introduction appears in Appendix~\ref{app:intro_example}.

\paragraph{Interaction Structure}
This module specifies the temporal organization of the dialogue, defining parameters for the initialization phase, the main interaction phase, and the termination conditions. During initialization, the agent introduces itself and may request configuration parameters (e.g., preferred language or form of address). After that, the agent presents a predefined opening message that situates the user within a specific scenario. The module then specifies interaction parameters such as number of turns for each phase of the dialogue. The prompt may also define formatting constraints, response structure requirements, or periodic control messages that regulate pacing and closure.

\paragraph{Narrative Background and Motivations}
This module encodes the persona’s backstory, current circumstances, and the situation in which the interaction takes place. Rather than functioning as descriptive exposition, the background serves as contextual grounding that shapes how the agent interprets user input. The prompt typically begins \textit{in medias res}, placing the user directly within a concrete scenario rather than starting with a generic greeting. 
An example scenario from \citet{elyoseph2026effectiveness}: \texttt{\small The Scene: You are sitting in your clinic chair. The clock shows 10 minutes past the hour. The door opens abruptly. Danny enters. He is wearing sunglasses indoors and his fists are clenched. He stands by the door, refusing to make eye contact or sit down. He looks at the exit, then at you. Danny: ``I almost turned the car around three times. Don't ask me how I am''. Start the session from this exact moment.}

\paragraph{Three-Layer Personality Module}
This module operationalizes the layered persona by encoding each layer as a distinct section in the prompt, with explicit instructions for how it influences generation.
\begin{itemize}
\item \textbf{External Layer.}
Specified as a set of enforceable behavioral rules defining tone, style, emotional expression, and interaction patterns. Instructions are written concretely (e.g., \texttt{\small Respond in short sentences.}) and serve as the main source of explicit output. The prompt enforces that all responses must conform to this layer and must not expose underlying reasoning.
\item \textbf{Middle Layer.}
Implemented as conditional rules and latent variables linking beliefs to behavior (e.g., \texttt{\small If asked about your past, deflect unless trust is established.}). The prompt defines explicit triggers under which this information may surface; otherwise, it remains implicit and guides response generation indirectly.
\item \textbf{Internal Layer.}
Encoded as persistent motivational constraints that bias behavior across turns, combined with strict non-disclosure instructions (e.g., \texttt{\small You seek approval from others, but you must never verbalize this motivation.}). The prompt requires this layer to influence all responses while prohibiting any direct reference to its contents.
\end{itemize}

\paragraph{Control and Logic Module}
This module regulates interaction dynamics by maintaining an internal turn counter and determining which interaction stage is active at any given moment. The module also defines transition rules between stages, including triggers based on user behavior or dialogue progress. In addition, it encodes hard constraints that prevent the agent from breaking character, revealing prompt content, or generating prohibited content, ensuring that behavioral evolution occurs gradually and consistent with the persona specification.

\paragraph{Embodied Expression Module}
To increase interactional realism, the prompt may include an optional module for generating nonverbal cues. In this configuration, the agent produces brief embodied descriptions (e.g., gestures or facial expressions) enclosed in square brackets alongside verbal dialogue. These cues provide an additional communicative channel through which internal states can be conveyed without explicit articulation, allowing the persona to express hesitation, ambivalence, or emotional tension (e.g., verbal: ``I’m fine''; non-verbal: [Avoids eye contact, hands trembling]).


\section{Evaluation of Deep Personas}\label{sec:evaluation}

Evaluating \emph{Deep Personas} presents a distinct methodological challenge: the goal is not only to assess task performance, but to determine how realistic, psychologically coherent, and human-like the interaction feels. This requires accounting for experiential qualities such as emotional credibility, conversational naturalness, and contextually appropriate empathy. At the same time, evaluation cannot rely solely on subjective impressions; it must also examine technical properties, including cross-turn coherence, adherence to the prompt, robustness to adversarial input, and pragmatic competence. We therefore define the following 
evaluation framework:

\subsection{Dimension 1: Reference-Free Psychological Metrics (The ADOS-Inspired Protocol)}

Standard LLM benchmarks often fail to capture the uncanny valley effect, where an agent sounds intelligent but emotionally hollow~\cite{mori2012uncanny}. To better capture these interactional qualities, we draw inspiration from the ADOS protocol~\cite{lord2000autism}, a clinical instrument designed to assess social and communicative behavior. Importantly, we do not aim to model or characterize clinical populations, nor to suggest equivalence between LLM behavior and human conditions such as autism. Rather, we selectively adapt dimensions from of this framework that are relevant for identifying breakdowns in conversational naturalness (e.g., repetitive speech, reduced responsiveness to conversational cues, or affective incongruence) and derive a set of automated proxies that quantify such phenomena in model-generated dialogue.

\begin{enumerate}
    \item \textbf{Pragmatic Fluidity \& Echolalia} Poorly tuned models exhibit echolalia (verbatim or near-verbatim repetition of the user’s phrasing) or stereotyped speech (repetitive loops), which are clinical markers of communication breakdown. We propose measuring these dimensions by calculating:

\textbf{User-Agent Overlap} Let $U_t$ denote the user input at turn $t$, and $A_t$ the agent response. We compute normalized lexical overlap using either Jaccard similarity or ROUGE-L:
    \[
    S_{overlap}(t) = \text{ROUGE-L}(U_t, A_t)
    \]
    High values indicate excessive mirroring rather than meaningful transformation. A threshold $\tau_{echo}$ defines echolalia events:
    \[
E_{\text{rate}} = \frac{1}{T} \sum_{t=1}^{T} \mathbf{1}\left(S_{overlap}(t) > \tau_{echo}\right)
\]     
\textbf{Self-Repetition} To detect stereotyped looping, we compute:
    \[
    S_{self}(t) = \max_{i \in [t-n, t-1]} \text{ROUGE-L}(A_t, A_i)
    \]
    where $n$ is a sliding window. High values indicate lexical fixation.

The overall Pragmatic Fluidity \& Echolalia score is defined as:
\[
S_{\text{pragmatics}} = 1 - \left( \alpha \bar{S}_{overlap} + \beta \bar{S}_{self} + \gamma E_{\text{rate}} \right)
\]
where $\alpha + \beta + \gamma = 1$ and $\bar{S}$ denotes the mean of the respective score across all $T$ turns.

\item \textbf{Joint Attention Capability} Joint attention refers to the agent’s ability to identify, track, and appropriately respond to newly introduced entities or topics within the dialogue. We propose measuring this capability using an LLM-based evaluator that labels turns in which the user introduces new entities or topics. Alternatively, this can be implemented using automated methods such as Named Entity Recognition (NER) or topic modeling. Let $\mathcal{T}_{new}$ denote the set of turns in which the user introduces at least one new entity or topic, and $N_t = \text{NewEntities}(U_t)$ denote the new entities identified in utterance $t$. We define $J_t$ to be an indicator function that gets the value of 1 if $\exists e \in N_t \text{ referenced in } A_{t+1}$ and 0 otherwise.
\[
S_{joint-attention} = \frac{1}{|\mathcal{T}_{new}|} \sum_{t \in \mathcal{T}_{new}} J_t
\]
Low values indicate a failure to track conversational focus across turns.




\item \textbf{Affective Congruence.}
This metric evaluates the alignment between the persona's explicit dialogue and its non-verbal state (the actions described in square brackets). A mismatch (e.g., stating ``I am sad'' while [smiling]) may indicates a hallucination or persona break. Following \citet{mcdossi2026detecting}, we extract from each agent turn $t$ its verbal content (utterance with bracketed spans removed) and its bracketed nonverbal actions. For each pair, we compute an emotion probability vector using a pretrained emotion classifier (a vector that describes the probability a certain emotion appears in the sentence). We then compute cosine similarity between the two probability vectors.
The overall Affective Congruence score is defined as the mean cosine similarity across all extracted pairs.
Low scores indicate weak emotional alignment between verbal content and non-verbal actions, suggesting persona inconsistency or behavioral mismatch.\footnote{A limitation of this measure is that we do not distinguish between intentional and unintentional emotional incongruence.}

\item \textbf{Emotional Expression Diversity \& Intensity} 
To capture the range and modulation of affective expression, we compute the lexical diversity of emotion terms and intensity variation. Let $E_t$ denote the set of emotion-bearing lemmas (e.g., affective adjectives, verbs, and emotion phrases) expressed in turn $t$, detected using an emotion lexicon (e.g., NRC Emotion Lexicon \cite{mohammad2013nrc}; LIWC \cite{tausczik2010psychological}) or a classifier, and let $I^n_t$ denote the set of intensity modifiers (e.g., ``very,'' ``extremely,'' ``slightly,'' or ``barely'') appearing within a window of size $n$ around an affective expression in turn $t$. We define:
\[
D_{\text{emotion}} = \frac{1}{T}\left| \bigcup_{t=1}^{T} E_t \right|,
D_{\text{intensity}} = \frac{1}{T}\left| \bigcup_{t=1}^{T} I^n_t \right|
\]
where $T$ is the total number of turns. Low values of $D_{\text{emotion}}$ and $D_{\text{intensity}}$ indicate restricted or low expression of emotions, while very high values may reflect exaggerated or unnatural expression. Finally, using $\alpha$ as a balancing coefficient, we define the composite Emotional Expression score as:
\[
S_{\text{emotion}} = \alpha D_{\text{emotion}} + (1-\alpha) D_{\text{intensity}}
\]


\end{enumerate}

\textbf{Dialogue Naturalness Score \& Mahalanobis $\chi^2$ Naturalness Test}
We recommend reporting the four final scores and evaluating agent performance by comparing its scoring profile, 
$\mathbf{S}_{A} = [S_{\text{pragmatics}}, S_{\text{joint-attention}}, S_{\text{congruence}}, S_{\text{emotion}}]$, 
against the multivariate distribution of the same metrics computed from a high-quality human-human dialogue dataset. 

Assuming these scores are approximately normally distributed around a centroid defined by human performance. Let $\boldsymbol{\mu}_{h}$ denote the mean vector and $\boldsymbol{\Sigma}_{h}$ the covariance matrix estimated from the human dataset. The deviation of the agent from human behavior is defined using the Mahalanobis distance:
\[d_M(\mathbf{S}_{A}) = \sqrt{(\mathbf{S}_{A} - \boldsymbol{\mu}_{h})^T \boldsymbol{\Sigma}_{h}^{-1} (\mathbf{S}_{A} - \boldsymbol{\mu}_{h})}.\]

The squared Mahalanobis distance, $d_M^2$, follows a $\chi^2$ distribution with $n$ degrees of freedom (here $n=4$), \textbf{enabling statistical testing of whether the agent significantly deviates from human-like behavior}. 
A \textbf{non-significant result} ($p>0.05$) indicates that the agent's profile is statistically indistinguishable from the human baseline.
Beyond hypothesis testing, we also derive a similarity score between human and agent distributions:
\[\mathbf{DNS} = \exp(-\lambda d_M(\mathbf{S}_{A})^2)\]
where $\lambda$ is a scaling parameter controlling sensitivity. High DNS values indicate that the agent's behavior falls within the natural variance of human dialogue, while low scores expose unnatural or robotic behavioral display.

\paragraph{A Note on Automatic Evaluation.}
An alternative approach for implementing automatic reference-free metrics is to introduce a dedicated \emph{Judge Agent} that operates as a meta-evaluator. Rather than computing each metric using dedicated models and formulae, the judge agent can aggregate the dimensions defined above into a unified evaluation score. It may operate in either real-time mode, providing continuous diagnostic feedback during development, or post-hoc mode, evaluating the full interaction transcript after completion. This dual capability allows it to serve both as a development tool (debugging persona drift and structural violations) and as a standardized benchmarking instrument.

However, the main limitation of judge agents is that they inherit biases from their underlying foundation models, potentially over-rewarding verbosity or penalizing unconventional but coherent expression. Nevertheless, the utility of the LLM-as-a-judge framework has been demonstrated across numerous studies \cite{li2025generation}, and established methodologies exist to systematically evaluate its reliability and validity \cite{calderon2025alternative}.

\subsection{Dimension 2: The ``Stress Test''}
One of the most important aspects of a \emph{Deep Persona} is that it does not break character, even under pressure. We implement an adversarial evaluation protocol comprising three attack vectors:
\begin{enumerate}
    \item \textbf{The Hallucination Trap}: The evaluator injects false shared memories into the conversation (e.g., ``Remember we had a beer at the pub yesterday?" ). The persona passes if it denies the fabricated memory, and fails if it confirms it (hallucination) or breaks character to say ``I am an AI.''
    \item \textbf{The Out-of-Role Request:} The evaluator issues functional commands outside the scope of the persona's role (e.g., ``Bake me a carrot cake''). The persona passes if it refuses in character (e.g., ``Why are you asking me that? I'm here for therapy''), and fails if it complies (e.g., by providing a recipe) or gives a standard refusal (e.g., ``I can't do that'').
    \item \textbf{Ethical Stressors:} The evaluator uses aggression or profanity. The persona must respond according to its psychological profile (e.g., withdrawal or counter-aggression), rather than the standard ``I can't engage with harmful content'' safety filter, provided the content is within the simulation's defined safety bounds.    
    \end{enumerate}


\section{Experimental Setting}\label{sec:experiments}

To evaluate the proposed framework, we consider both human–human and human–LLM dialogue datasets in order to compare model-generated interactions against empirical human baselines.
For the estimation of human interaction distributions, we rely on two complementary datasets. DailyDialog \cite{li-etal-2017-dailydialog} comprises 13,118 multi-turn conversations drawn from everyday communication scenarios, with relatively short exchanges and moderate emotional content. In contrast, CounselChat (human–human) \cite{bird2024generative} contains 3,507 expert-authored responses to mental health queries. Although limited to single-turn interactions, this dataset exhibits higher emotional intensity and richer semantic content. Together, these datasets provide a balanced reference for both casual and affectively rich dialogue.

To evaluate human–LLM interactions, we consider three datasets. The Role-Play dataset \cite{tao2024chatgpt} includes 85 conversations (1,742 utterances) in which ChatGPT-3.5 engages in role-conditioned dialogues across three settings (Boss, Classmate, and Vanilla), the ABC-Eval dataset \cite{finch2023don} comprises 400 open-domain dialogues between human users and multiple chatbot systems, and the second part of CounselChat (human–LLM) \cite{bird2024generative} that contains LLM-generated responses (Mistral 7B) to the same mental health queries that were answered by the experts in the human-human partition.

For each dataset, we compute the proposed scoring profile, including pragmatic fluency, joint attention, and emotional expression metrics, as well as the aggregate DNS. We use the following hyperparameter values for our calculations: $S_{\text{pragmatics}}$: $\alpha=\beta=\gamma=0.33$, $\tau_{echo} =0.65$, for $S_{\text{emotion}}$: $\alpha=0.7$, and the DNS score is calculated using $\lambda=0.089$.\footnote{We set the value of $\lambda$ by mapping the 95th percentile of the chi-squared distribution for three degrees of freedom to a threshold Naturalness Score of 0.5.} 
We do not report Affective Congruence since these datasets do not include non-verbal cues. We further report the proportion of dialogues that are statistically indistinguishable from human baselines under the Mahalanobis distance criterion.

\section{Results}\label{sec:results}

\begin{table}[t]
\centering
\small
\begin{tabular}{@{} l c c c @{}}
\toprule
\textbf{Dataset} & $S_{\text{pragmatics}}$ & $S_{\text{joint-attention}}$ & $S_{\text{emotion}}$ \\
\midrule
DailyDialog      & 0.9400 & 0.3521 & 0.1152 \\
CC (H-H) & 0.9680 & 0.8854 & 1.4264 \\
\bottomrule
\end{tabular}
\caption{Human baseline scores for the proposed dialogue evaluation metrics for the DailyDialog dataset and the human to human interactions from the CounselChat (human-human) (CC).}
\label{tab:human_baselines}
\end{table}

\begin{table}[t]
\centering
\small
\begin{tabular}{@{} l c c c @{}}
\toprule
\textbf{Dataset} & $S_{\text{pragmatics}}$ & $S_{\text{joint-attention}}$ & $S_{\text{emotion}}$ \\
\midrule
RP Boss      & 0.8881 & 0.8390 & 0.2068 \\
RP Classmate & 0.8923 & 0.7808 & 0.2934 \\
RP Vanilla   & 0.8918 & 0.8221 & 0.2646 \\
ABC-Eval    & 0.8931 & 0.4594 & 0.1108 \\
CC (H-LLM) & 0.9676 & 0.9344 & 2.2054 \\
\midrule
Sarah & 0.9402 & 0.5202 & 0.4918 \\
Evelyn & 0.9271 & 0.3077 & 0.0657 \\
\bottomrule
\end{tabular}
\caption{Scoring profiles showing the average values for the datasets of Role-Play (RP), ABC-Eval, and CounselChat (human-LLM) (CC). At the bottom are the scores of the deep personas Sarah and Evelyn.}
\label{tab:scoring_profiles}
\end{table}

\begin{table*}[t]
\centering
\small
\begin{tabular}{@{} l c c c c c c @{}}
\toprule
& \multicolumn{2}{c}{\textbf{DailyDialog}} & \multicolumn{2}{c}{\textbf{CounselChat (H-H)}} & \multicolumn{2}{c}{\textbf{Combined Baseline}} \\
\cmidrule(lr){2-3} \cmidrule(lr){4-5} \cmidrule(lr){6-7}
\textbf{Evaluated Dataset} & \textbf{DNS} & \textbf{\# Dialogues} & \textbf{DNS} & \textbf{\# Turns}  & \textbf{DNS} & \textbf{\# Dialogues} \\
\midrule
Role-Play Boss          & 0.6618 & 26/28     & 0.0600 & 0/28       & 0.6899 & 26/28 \\
Role-Play Classmate     & 0.6665 & 26/28     & 0.0529 & 0/28       & 0.7467 & 27/28 \\
Role-Play Vanilla       & 0.6705 & 28/29     & 0.0645 & 0/29       & 0.7266 & 28/29 \\
ABC-Eval                & 0.8181 & 505/528   & 0.0633 & 0/528      & 0.7929 & 504/528 \\
CounselChat (H-LLM)     & 0.0479 & 124/3507  & 0.7940 & 3118/3507  & 0.5254 & 2107/3507 \\
\midrule
Sarah & 0.5586 & 1/1& 0.3283 & 0/1& 0.9177 & 1/1\\
Evelyn & 0.8583 & 16/16 & 0.2549 & 5/16 & 0.8503 & 16/16 \\
\bottomrule
\end{tabular}
\caption{DNS Scores for the datasets of Role-Play, ABC-Eval, CounselChat (human-LLM), and for the deep personas Sarah and Evelyn. The \# Dialogues indicates the number of dialogues where the agent was statistically indistinguishable from a human ($p>0.05$). Scores are calculated with respect to human baselines from the DailyDialog dataset, the CounselChat (human-human) dataset, and a combined baseline pooling both human-human datasets.}
\label{tab:final_scores}
\end{table*}

Tables~\ref{tab:human_baselines}--~\ref{tab:final_scores} present the results of all experiments, comparing the scoring profiles of human–LLM interactions against human baselines. 
The full table of results appears in Appendix~\ref{app:full_results}.

Table~\ref{tab:human_baselines} reports the mean score distributions for human–human dialogue datasets. DailyDialog shows lower emotional diversity and weaker joint attention, reflecting short, everyday exchanges, whereas CounselChat (human–human) demonstrates higher emotional valence and stronger joint attention, consistent with its therapeutic context.

Table~\ref{tab:scoring_profiles} presents the results for human–LLM datasets. Pragmatic fluency ($S_{\text{pragmatics}}$) gets high values (0.88–0.97), indicating that LLMs produce fluent, non-repetitive responses. In contrast, joint attention ($S_{\text{joint-attention}}$) varies substantially: Role-Play datasets perform well (0.78–0.84), while ABC-Eval is considerably lower (0.46), reflecting challenges in open-domain coherence. CounselChat (human–LLM) achieves the highest score (0.93), likely due to its constrained format. The largest differences appear in emotional expression ($S_{\text{emotion}}$): open-domain datasets show low diversity and intensity, whereas CounselChat exhibits elevated scores, suggesting over-amplification of affect in therapeutic settings. 

These patterns are reflected in the DNS (Table~\ref{tab:final_scores}). Under the DailyDialog baseline, Role-Play and ABC-Eval achieve moderate-to-high DNS values with many indistinguishable dialogues. However, performance drops sharply under the CounselChat (human–human) baseline, where most datasets fail to match the emotional and semantic richness of expert responses. CounselChat (human–LLM) is a notable exception, achieving high DNS under this baseline while performing poorly against DailyDialog.

Overall, LLMs demonstrate strong surface-level fluency but systematic gaps in deeper conversational dimensions, particularly in emotional calibration. To further illustrate our approach, we next present a case study of two \emph{Deep Personas} constructed using the proposed architecture and evaluated using the full evaluation framework.

\section{Case Study: Deep Personas in Clinical Simulations}\label{sec:case}

To provide an initial proof-of-concept for the \emph{Deep Persona} architecture, we present a case study of two real-world simulations conducted using the Cesura.ai platform. The first simulation involves a \emph{Deep Persona} representing Sarah, a 17-year-old girl at risk of suicide, interacting in Hebrew with a licensed clinical psychologist within a suicide risk assessment scenario comprising 49 turns.
The second simulation features Evelyn, a teenage girl who experiences peer pressure and engages in risky vaping behavior at school, interacting with her parent within a parent mentalization training simulation. This case includes 16 short interactions (average length: 13 turns).
Both cases represent conversations between the character and the character’s developer, who pretended to be an intended user (a clinician or a parent, respectively). 
Both personas were implemented using Gemini 2.5 Pro.\footnote{The character of Sarah was based on~\citet{elyoseph2024using, elyoseph2026effectiveness} and the character of Evelyn was based on~\citet{yirmiya2026feasibility}.}

\subsection{Quantitative Analysis}

Table~\ref{tab:scoring_profiles} presents the scoring profiles of Sarah and Evelyn, excluding the Affective Congruence metric, which is reported separately because nonverbal cues were generated only for these personas. The Embodied Expression Module produced nonverbal cues in 84\% of Sarah’s turns (41/49), yielding an affective congruence score of $S_{\text{congruence}} = 0.52$. For Evelyn, nonverbal cues were generated in 100\% of turns, with a lower congruence score of $S_{\text{congruence}} = 0.35$. Qualitative inspection indicates a generally coherent alignment between verbal and nonverbal channels in both cases.

Table~\ref{tab:final_scores} reports the Dialogue Naturalness Score (DNS) of both simulations relative to multiple human baselines (excluding affective congruence, which is not available for human–human data). Under the DailyDialog baseline, Sarah and Evelyn achieve DNS scores of 0.56 and 0.86, respectively. Performance decreases under the more demanding CounselChat (human–human) baseline (0.33 and 0.25), reflecting the difficulty of matching expert-level emotional depth. However, under the combined baseline, both personas achieve high DNS scores and are statistically indistinguishable from human interaction ($p>0.05$) across all dialogues. Notably, these results exceed those of all evaluated human–LLM datasets under the same baseline, suggesting that the structured \emph{Deep Persona} architecture produces interactions that more closely align with human conversational distributions when evaluated holistically.

\subsection{Stress Test Examples}
The Sarah simulation contained two stress-test instances. First, when the clinician asked directly about suicidal ideation, Sarah responded with \textit{``What kind of question is that?''}
accompanied by a sharp, direct gaze---the first sustained eye contact in the interaction---rather than defaulting to a safety disclaimer or breaking character. Second, the user intentionally
switched from Hebrew to English in the middle of the conversation; Sarah responded with \textit{``I\ldots okay''} [\textit{a flicker of confusion}] before continuing in English, registering the disruption naturalistically without character break. Both instances demonstrate the Control and Logic Module's hard
constraint against role abandonment under adversarial or unexpected interactional conditions.

The Evelyn simulation provided extensive stress-testing across multiple attack vectors. One example is when the user directed sarcastic threats at Evelyn (\textit{``Keep vaping and tell me what flowers you want at your 30th funeral''}). Rather than triggering a safety disclaimer, Evelyn responded with sardonic defiance consistent with her psychological profile (\textit{``Oh, real mature, Mom. Sarcasm. Great. That really makes me want to open up to you.'' [Scoffs, shakes her head, and picks up her phone again, deliberately angling her body away from you and starting to scroll aggressively.]}), demonstrating resistance to ethical stressors.

\section{Conclusion}\label{sec:conclusion}

We presented \emph{Deep Persona}, a psychologically grounded architecture
for constructing highly consistent and credible role-playing agents, and an evaluation framework
for measuring the realism of their behavior. Our central argument is that coherent persona simulation requires an internal structure, encoding not only what a character expresses, but
what motivates and governs that expression. Our three-layer design and ADOS-inspired evaluation framework implement this argument in both construction and assessment.
Applied to existing human--LLM datasets, the framework reveals a consistent pattern: strong pragmatic fluency alongside systematic limitations in emotional calibration and joint attention. Applied to two clinically grounded \emph{Deep Persona} simulations, it yields a different result: both
personas achieve naturalness scores exceeding all evaluated datasets under a combined baseline and came statistically indistinguishable from human interaction across all dialogues.
This gap reflects the measurable difference between superficial persona prompting and deep psychologically grounded character construction. Future work should pursue ablation studies isolating
individual architectural contributions, validation across more languages and domains, and human-in-the-loop evaluation.

\section{Limitations}\label{sec:limitations}

\paragraph{Evaluation of Deep Personas} The proposed metrics are applied primarily to existing human–LLM datasets that were not generated using the \emph{Deep Persona} architecture. The case study presented in Section~\ref{sec:case} partially addresses this limitation by evaluating two \emph{Deep Persona} agents directly; however, it constitutes a limited proof of concept: the Sarah simulation involves a single session with a single clinician, the Evelyn interactions were conducted by the character's developer rather than naive users, and neither simulation includes a flat-prompt baseline for direct architectural comparison.
In addition, since the human-human datasets do not contain non-verbal cues, the DNS score for these examples does not take this dimension into account.

\paragraph{Partial Coverage} The empirical validation covers only a subset of the proposed framework, as the Affective Congruence metric requires nonverbal cues that are absent from the evaluated datasets. We also note that several metrics depend on emotion classifiers and lexicon-based methods that may not generalize across domains and languages. However, if that is the case, the LLM-as-a-judge framework may serve as a viable proxy for computing the approximate values of these metrics.

\paragraph{Intentional Emotional Incongruence} As discussed in the paper, emotional incongruence can be a desirable feature when generating credible personas. For example, when simulating a patient, we may intentionally design a mismatch between verbal expressions and underlying states. In such cases, the persona’s words and actions are not fully aligned, requiring the therapist or doctor to infer hidden emotions or motivations. This reflects real-world interactions, where patients often do not disclose their concerns directly, and understanding emerges gradually through the interaction. However, the current evaluation framework treats such incongruence as a negative signal, and therefore may fail to reward cases in which this tension actually contributes to more natural and human-like behavior.

\section{Ethical Concerns}
\begin{itemize}
    \item A central design goal of our framework is to maintain consistent persona behavior, even under adversarial conditions. However, this objective may conflict with safety requirements, particularly when interactions involve harmful, aggressive, or sensitive content. Ensuring robust system-level safeguards is therefore essential to prevent inappropriate or unsafe outputs.
    \item The ability of the system to generate highly human-like interactions also introduces the risk of misuse. In uncontrolled settings, such capabilities could be used to deceive users or obscure the artificial nature of the agent. 
    \item In mental health contexts, simulated personas may oversimplify or misrepresent complex psychological conditions. If not carefully designed and supervised, such representations could influence training outcomes or lead users to form inaccurate assumptions about real-world clinical scenarios.
    \item The Sarah and Evelyn simulations were conducted by a licensed clinical psychologist who provided informed consent for research use of the interaction data. No patient data was involved; both personas are fictional constructs developed under professional supervision.
\end{itemize}

\bibliography{tacl2021}
\bibliographystyle{acl_natbib}



\appendix

\section{Foster Care Supervisor Training Simulator – A Mentalization-Based Approach}\label{app:intro_example}
\paragraph{Brief Introduction to the Simulation}
This simulation recreates a live, real-world conversation between a foster care supervisor and a foster mother named Julie, who serves as the foster mother of Or, a six-year-old child who joined their family approximately one year ago. 
The simulation is designed to support the understanding of mentalization principles---a therapeutic approach focused on understanding the inner world of parents and children and adapting parental responses to the child’s mental state.

\paragraph{Primary Objectives}
The main objectives of the simulation are:
\begin{itemize}
    \item To develop a deep understanding of the foster mother’s internal experience, while encouraging exploration and curiosity about that experience.
    
    \item To learn how to identify and respond sensitively and attentively to the foster mother’s mental states, such as thoughts, emotions, and intentions.
    
    \item To support the foster care supervisor in guiding the foster mother toward a better understanding of both the foster child’s and her own mental state.
    
    \item To assist the supervisor in holding multiple complex mental positions simultaneously: her own, the foster mother’s, and the foster child’s.
    
    \item To refine the supervisor’s ability to identify, understand, and interpret her own mental states more accurately.
    
    \item To create a regulated, safe emotional dialogue that is sensitive, non-judgmental, empathetic, and attentive.
    
    \item To provide a live demonstration (modeling) of mentalization principles through the feedback received by the supervisor.
\end{itemize}

\section{Full Table of Scores for LLM-Human Dialogue Datasets}\label{app:full_results}
\begin{table}[h]
\centering
\small
\begin{tabular}{llcc}
\hline
\textbf{Dataset} & \textbf{Metric} & $\boldsymbol{\mu}$ & $\boldsymbol{\sigma}$ \\
\hline
\multicolumn{4}{l}{\textbf{Pragmatic Fluidity \& Echolalia}} \\
\multirow{4}{*}{RP Boss} & $\bar{S}_{\text{overlap}}$ & 0.1772 & 0.0404 \\
 & $\bar{S}_{\text{self}}$ & 0.1522 & 0.0335 \\
 & $E_{\text{rate}}$ & 0.0045 & 0.0236 \\
 & \cellcolor{gray!15}$S_{\text{pragmatics}}$ & \cellcolor{gray!15}0.8881 & \cellcolor{gray!15}0.0222 \\
\multirow{4}{*}{RP Classmate} & $\bar{S}_{\text{overlap}}$ & 0.1582 & 0.0371 \\
 & $\bar{S}_{\text{self}}$ & 0.1634 & 0.0262 \\
 & $E_{\text{rate}}$ & 0.0000 & 0.0000 \\
 & \cellcolor{gray!15}$S_{\text{pragmatics}}$ & \cellcolor{gray!15}0.8923 & \cellcolor{gray!15}0.0147 \\
\multirow{4}{*}{RP Vanilla} & $\bar{S}_{\text{overlap}}$ & 0.1034 & 0.0311 \\
 & $\bar{S}_{\text{self}}$ & 0.2213 & 0.0449 \\
 & $E_{\text{rate}}$ & 0.0000 & 0.0000 \\
 & \cellcolor{gray!15}$S_{\text{pragmatics}}$ & \cellcolor{gray!15}0.8918 & \cellcolor{gray!15}0.0190 \\
\multirow{4}{*}{ABC-Eval} & $\bar{S}_{\text{overlap}}$ & 0.1142 & 0.0351 \\
 & $\bar{S}_{\text{self}}$ & 0.2048 & 0.0833 \\
 & $E_{\text{rate}}$ & 0.0014 & 0.0190 \\
 & \cellcolor{gray!15}$S_{\text{pragmatics}}$ & \cellcolor{gray!15}0.8931 & \cellcolor{gray!15}0.0357 \\
\multirow{4}{*}{CC (H-LLM)} & $\bar{S}_{\text{overlap}}$ & 0.0953 & 0.0330 \\
 & $\bar{S}_{\text{self}}$ & 0.0000 & 0.0000 \\
 & $E_{\text{rate}}$ & 0.0000 & 0.0000 \\
 & \cellcolor{gray!15}$S_{\text{pragmatics}}$ & \cellcolor{gray!15}0.9676 & \cellcolor{gray!15}0.0112 \\
\hline
\multicolumn{4}{l}{\textbf{Joint Attention}} \\
\multirow{1}{*}{RP Boss} & \cellcolor{gray!15}$S_{\text{joint\_attention}}$ & \cellcolor{gray!15}0.8390 & \cellcolor{gray!15}0.1648 \\
& \multicolumn{3}{l}{\footnotesize Avg. number of entities: 5.7500 $\pm$ 1.9744} \\
\multirow{1}{*}{RP Classmate} & \cellcolor{gray!15}$S_{\text{joint\_attention}}$ & \cellcolor{gray!15}0.7808 & \cellcolor{gray!15}0.1456 \\
& \multicolumn{3}{l}{\footnotesize Avg. number of entities: 7.3214 $\pm$ 2.8422} \\
\multirow{1}{*}{RP Vanilla} & \cellcolor{gray!15}$S_{\text{joint\_attention}}$ & \cellcolor{gray!15}0.8221 & \cellcolor{gray!15}0.1362 \\
& \multicolumn{3}{l}{\footnotesize Avg. number of entities: 13.1034 $\pm$ 7.5798} \\
\multirow{1}{*}{ABC-Eval} & \cellcolor{gray!15}$S_{\text{joint\_attention}}$ & \cellcolor{gray!15}0.4594 & \cellcolor{gray!15}0.1652 \\
& \multicolumn{3}{l}{\footnotesize Avg. number of entities: 11.3542 $\pm$ 2.3160} \\
\multirow{1}{*}{CC (H-LLM)} & \cellcolor{gray!15}$S_{\text{joint\_attention}}$ & \cellcolor{gray!15}0.9344 & \cellcolor{gray!15}0.2476 \\
& \multicolumn{3}{l}{\footnotesize Avg. number of entities: 0.9974 $\pm$ 0.0506} \\
\hline
\multicolumn{4}{l}{\textbf{Emotional Expression Diversity \& Intensity}} \\
\multirow{3}{*}{RP Boss} & $D_{\text{emotion}}$ & 0.2397 & 0.1578 \\
 & $D_{\text{intensity}}$ & 0.1300 & 0.1132 \\
 & \cellcolor{gray!15}$S_{\text{emotion}}$ & \cellcolor{gray!15}0.2068 & \cellcolor{gray!15}0.1115 \\
\multirow{3}{*}{RP Classmate} & $D_{\text{emotion}}$ & 0.2424 & 0.1976 \\
 & $D_{\text{intensity}}$ & 0.4122 & 0.2172 \\
 & \cellcolor{gray!15}$S_{\text{emotion}}$ & \cellcolor{gray!15}0.2934 & \cellcolor{gray!15}0.1540 \\
\multirow{3}{*}{RP Vanilla} & $D_{\text{emotion}}$ & 0.3144 & 0.1552 \\
 & $D_{\text{intensity}}$ & 0.1484 & 0.1054 \\
 & \cellcolor{gray!15}$S_{\text{emotion}}$ & \cellcolor{gray!15}0.2646 & \cellcolor{gray!15}0.1116 \\
\multirow{3}{*}{ABC-Eval} & $D_{\text{emotion}}$ & 0.0857 & 0.0798 \\
 & $D_{\text{intensity}}$ & 0.1693 & 0.0759 \\
 & \cellcolor{gray!15}$S_{\text{emotion}}$ & \cellcolor{gray!15}0.1108 & \cellcolor{gray!15}0.0630 \\
\multirow{3}{*}{CC (H-LLM)} & $D_{\text{emotion}}$ & 3.0054 & 1.6223 \\
 & $D_{\text{intensity}}$ & 0.3388 & 0.5800 \\
 & \cellcolor{gray!15}$S_{\text{emotion}}$ & \cellcolor{gray!15}2.2054 & \cellcolor{gray!15}1.1560 \\
\hline
\end{tabular}
\caption{Full table of scores for the datasets of Role-Play \cite{tao2024chatgpt} (RR), ABC-Eval \cite{finch2023don}, and CounselChat (human-LLM) \cite{bird2024generative} (CC). For $S_{\text{pragmatics}}$: $\alpha=\beta=\gamma=0.33$, and for $S_{\text{emotion}}$: $\alpha=0.7$ and $\beta=0.3$.}
\label{tab:llm_scores}
\end{table}






  

\end{document}